\documentclass[letterpaper, 10 pt, conference]{ieeeconf}
\let\labelindent\relax
\usepackage{enumitem}
\usepackage{graphicx}
\usepackage{xcolor}
\usepackage{tikz}
\usepackage{tikzscale}
\usepackage{tensor}
\usepackage{cite}
\usetikzlibrary{bayesnet}
\usepackage{amsmath}
\usepackage{mathtools}
\usepackage{amsfonts}
\usepackage{bbold}
\usepackage{tabularx}
\usepackage{courier}
\usepackage{balance}
\usepackage{hyperref}

\IEEEoverridecommandlockouts % you want to use the \thanks command
\newcommand{\mv}[1]{\mathbf{#1}}
\newcommand{\ctrans}[3]{\tensor*[^{#1}]{\mv{#2}}{_{#3}}}
\newcommand{\ctransi}[3]{\tensor*[^{#1}]{\mv{#2}}{^{-1}_{#3}}}
\newcommand{\cvec}[2]{\tensor*[^{#1}]{\mv{#2}}{}}
\DeclareMathOperator*{\argmax}{arg\,max}
\begin{document}

\title{\LARGE \bf TagSLAM: Robust SLAM with Fiducial Markers}
\author{
Bernd Pfrommer$^{1,2}$\\
\and
Kostas Daniilidis$^{1}$\thanks{}
\thanks{$^{1}$University of Pennsylvania School of Engineering and Applied
Science. We gratefully acknowledge support through the following grants: NSF-IIP-1439681
(I/UCRC RoSeHuB), NSF-IIS-1703319, NSF MRI 1626008, and the DARPA FLA
program.}
\thanks{$^{2}$Thanks to Chao Qu and Ke Sun for many useful discussions.}
}

\maketitle
\begin{abstract}
TagSLAM provides a convenient, flexible, and robust way of performing
Simultaneous Localization and Mapping (SLAM) with AprilTag fiducial
markers. By leveraging a few simple abstractions (bodies, tags,
cameras), TagSLAM provides a front end to the GTSAM factor graph
optimizer that makes it possible to rapidly design a range of
experiments that are based on tags: full SLAM, extrinsic camera
calibration with non-overlapping views, visual localization for ground
truth, loop closure for odometry, pose estimation etc. We discuss in
detail how TagSLAM initializes the factor graph in a robust way, and
present loop closure as an application example.  TagSLAM is a ROS
based open source package and can be found
at \url{https://berndpfrommer.github.io/tagslam_web}.
\end{abstract}
\section{Introduction}
\begin{figure}[ht]
  \centering
  \includegraphics[width=0.8\columnwidth]{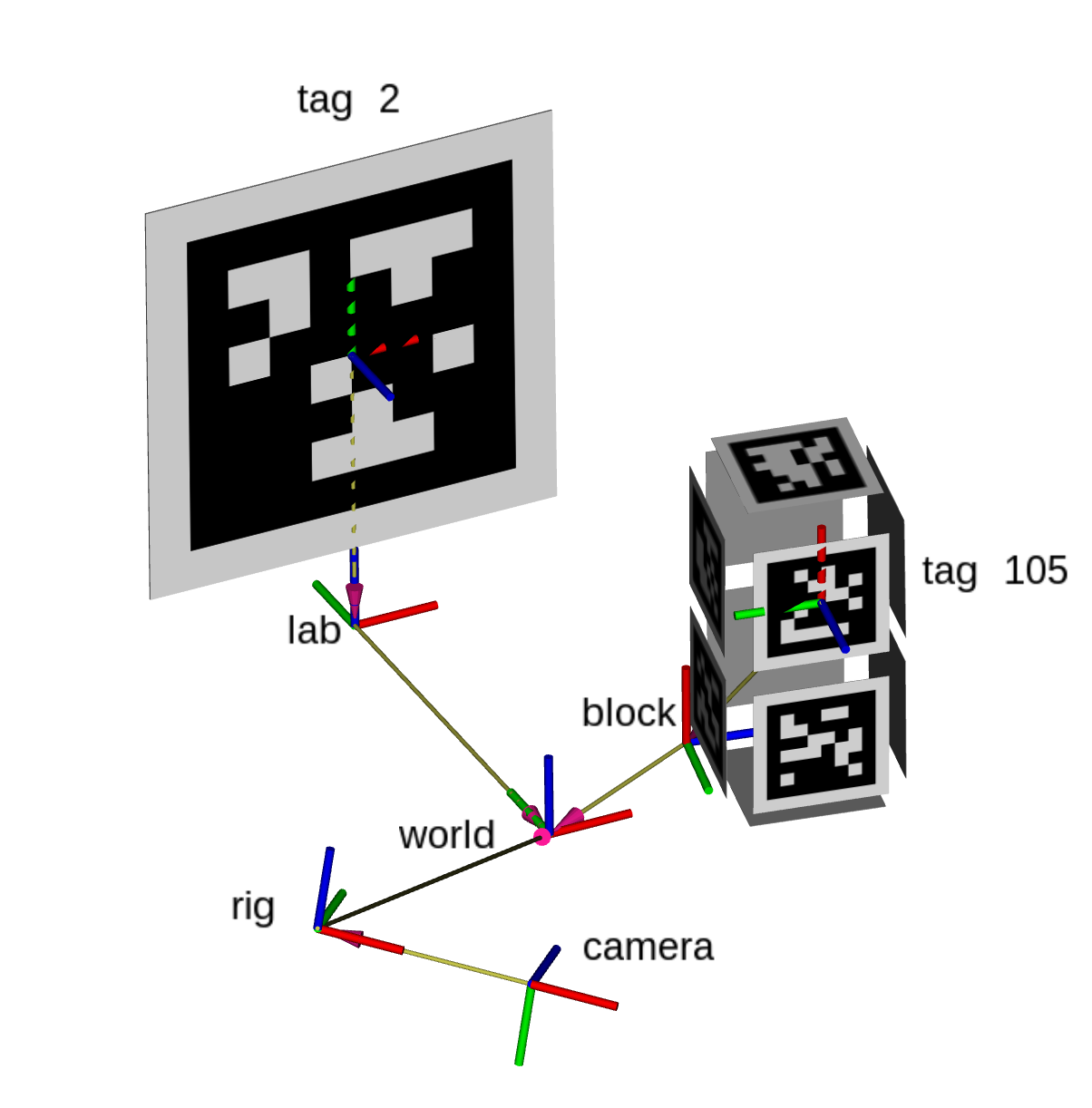}
  \caption{Single-camera TagSLAM scene with dynamic ``rig'' and ``block''
    bodies, and a static ``lab'' body.}
  \label{fig:scene_with_block}
\end{figure}

Accurate and robust Simultaneous Localization And Mapping (SLAM)
algorithms are arguably one of the corner pieces for building
autonomous systems. For this reason, SLAM has been studied extensively
since the 1980s \cite{cadena2016}. 
In a nutshell, SLAM methods take sensor data such as camera images as
input and extract easily recognizable features (``landmarks''). The
landmarks are entered into a map such that when the same landmark is
detected again later, the observer's pose can be determined by
triangulation. In graph based SLAM, a nonlinear optimizer such as
GTSAM \cite{kaess2011} is used to simultaneously optimize the pose of
the observer and the location of the landmarks, given the sensor
measurements.

Much research effort \cite{cadena2016} has gone into solving the
hard aspects of SLAM. One of them is recognizing previously seen
landmarks (''loop closure''). This is often difficult if landmarks are
observed from a different viewpoint, or under different lighting
conditions. The other is maintaining a map of landmarks. To limit memory 
consumption and achieve fast retrieval, landmarks must be at some
point discarded, efficient retrieval databases must be updated and
queried, adding considerable complexity \cite{murartal2017}. If loop
closure fails, the estimated camera pose starts to drift, which is a
well-known problem with local-map-only algorithms such as
visual-inertial odometry (VIO).

In many situations, and in particular for laboratory experiments these
problems can be sidestepped with a mild intervention to the 
environment. By placing visual fiducial markers such as the
popular AprilTags \cite{wang2016} in a scene, one can create a small
set of artifical landmarks, typically numbering less than 100. The
TagSLAM framework presented in this paper is designed for such a situation.
With only a few landmarks to consider, memory and retrieval speed are
no longer critical, and loop closure is 
guaranteed once a tag is successfully decoded by the AprilTag
library. If one or more tags are visible at all times, then TagSLAM
can perform both tracking and loop closure.

But more importantly, tags permit the introduction of geometric
constraints that have been observed by other means. For instance, markers
can be placed inside a building at coordinates that are known because
a highly accurate building plan is available
\cite{pfrommer2017}. Another alternative is measuring the locations
of select tags with a laser distance measuring tool. When a measured
marker is later recognized by the moving camera, it will pull the
trajectory to the reference location. As we demonstrate in the
experimental section, this can augment existing SLAM or
VIO methods. By utilizing well-known tags in a few places, and
ubiquitous but anonymous feature points elsewhere, one can leverage
the advantages of both approaches.

Even though tag based SLAM is much simpler than general SLAM, some
remaining difficulties are addressed in this paper. Since TagSLAM
utilizes GTSAM  \cite{kaess2011}, a graph based non-convex optimizer, care must be taken
to properly initialize all starting values. A bad initial value can
cause the optimizer to fail or to converge to a local minimum. In the
present work we give a detailed description of how we achieve robust
initialization to make TagSLAM a tool that can be applied to a wide range of
real-world situations without parameter tuning.

It is worthwhile noting that SLAM is a rather general algorithm, and
therefore can solve several related, simpler problems as well. If for
example the camera pose is known, one can estimate the pose of the
landmarks. Likewise, if a map of the landmarks is known, the camera
pose can be inferred. TagSLAM inherits all of this flexibility. In
contrast to traditional SLAM packages however, where moving landmarks
are filtered out, TagSLAM can explicitly model and track multiple
objects that have tags attached to them. This means TagSLAM can also
be used for pose estimation of objects other than cameras, as shown in
Fig. \ref{fig:scene_with_block}. In fact, TagSLAM has been deployed for
the University of Pennsylvania's Smart Aviary project in a scenario
where {\em all} cameras are static, and only some tags are
moving. This is not directly possible with any SLAM package that we
are aware of.

TagSLAM's robustness and
availability as an open-source ROS \cite{quigley2009} package make it
particularly accessible for researchers new to robotics. It is now
being used at the University of Pennsylvania's GRASP lab for extrinsic
camera calibration with non-overlapping views, loop closure on
visual-inertial odometry benchmarks, tag mapping, object pose
estimation, and more. TagSLAM is also employed in a forthcoming
project by the University of Michigan's DASC laboratory.

\section{Related Work}
As a review of the vast SLAM literature is beyond the scope of this
paper, we refer to the thorough survey in Ref.\ \cite{cadena2016}. In
this section we will focus on SLAM that utilizes fiducial markers.

Ref. \cite{neunert2016} presents a visual-inertial approach for
obtaining ground truth positions from a combination of inertial
measurement unit (IMU) and camera. They also utilize April\-Tags, but in
contrast to TagSLAM use an Extended Kalman Filter (EKF), a method that
is well suited for their goal to effectively provide a real-time,
outdoor, low-cost motion capture system. Since TagSLAM uses a graph
optimizer instead of a filter, it can leverage observations from both
the past and the future, resulting in 
smoother trajectories. In case IMU data is available, TagSLAM can be set up
to operate in a similar fashion to the system described in
\cite{neunert2016} by supplying externally generated VIO data.

Most similar to our work is Ref.\ \cite{munoz2018}, where a system is
described that can perform real-time SLAM from square planar markers
(SPM-SLAM). By using keyframes and having different algorithms for
tracking and loop closure, they achieve high performance and
scalability to large scenes. In
contrast, TagSLAM considers every frame a keyframe, and relies on
iSAM2 \cite{kaess2011} to exploit sparsity. This significantly simplifies
TagSLAM's code base, and facilitates the implementation of its
flexible model.
For long trajectories and many tags, we expect SPM-SLAM to scale
better due to its use of keyframes. However, our
experiments (Sec. \ref{sec:application}) indicate that TagSLAM can handle
scenes that are sufficiently large for many situations. Unlike
TagSLAM, SPM-SLAM cannot incorporate keypoints via odometry
measurements, although this shortcoming is addressed in a forthcoming
paper (``UcoSLAM''). As far as robustness is concerned, both TagSLAM and
SPM-SLAM rely on detecting tag pose ambiguity \cite{schweighofer2006},
but SPM-SLAM attempts to use the measurements right away by relying on
a sophisticated two-frame initialization algorithm, whereas TagSLAM delays use
of the tag's observations until its pose can be determined
unambiguously. For achieving robustness during relocalization,
SPM-SLAM relies again on pose ambiguity measures, whereas TagSLAM
additionally considers the tag's apparent size.

Also closely related to the present paper is the underwater SLAM performed
in \cite{westman2018}. The authors leverage AprilTags and the
GTSAM factor graph optimizer to obtain vision based ground truth poses
and extrinsic calibration. The focus of their work however is more on
providing a solution for a particular problem, whereas TagSLAM aims to
be a general purpose framework that can be easily applied to many
different settings. In fact, all the factors in \cite{westman2018}
that are related to AprilTags are already implemented in TagSLAM. The
code structure of TagSLAM is designed such that adding the
problem-specific XHY and ZPR factors from \cite{westman2018} should be
straight forward.

\section{Model Setup}
TagSLAM uses a few simple abstractions with which complex scenarios
can be built without the need to write any code. This section will
introduce the concepts and establish the necessary notation.

We denote as $\ctrans{B}{T}{A}$ an SE(3) transform that takes vector
coordinates $\cvec{A}{X}$ in reference system $A$ and expresses them in $B$:
\begin{equation}
\cvec{B}{X} = \ctrans{B}{T}{A}\cvec{A}{X}.
\end{equation}
Such a transform defines a pose. We distinguish two kinds:
\begin{itemize}
  \item Static pose: the transform $\ctrans{B}{T}{A}$ is independent
    of time. A static poses is represented by a single variable for
    the optimization process. Note that this does not mean the pose
    must be known from the beginning (i.e have a prior), but could be
    discovered as image data becomes available.
  \item Dynamic pose: the transform $\ctrans{B}{T}{A}(t)$ is time
    dependent. Such a pose is assumed to change, i.e. the optimizer
    will allocate a new variable for every time step $t$. Sufficient
    input data must be provided at time $t$ to solve for
    $\ctrans{B}{T}{A}(t)$.
\end{itemize}

A {\em body} is an object that can have tags and cameras attached to
it. Its pose is always given with respect to world coordinates, and
can be classified as static or dynamic depending on the nature of the
body. For static bodies, an optional prior pose may be specified.

Tags must have a unique id, and each tag must be associated with a
body to which it is attached. Tags without association are ignored,
unless a default body is specified to which any unknown tag will be
attached upon discovery.  In contrast to body poses, tag poses are
{\em always static}, and are expressed with respect to the pertaining
body. A tag pose prior is optional, so long as that is not required to
determine the body pose.

Camera poses, like tag poses, are {\em static}, and given with respect
to the body to which the camera is attached. This body is referred to
as the camera's ``rig'', although from a modeling point of view, it is
a body like any other, and can have for instance tags attached to
it. A prior camera pose (extrinsic calibration) is optional provided
the optimization problem can be solved without it.

With bodies, tags, and cameras, a rich set of SLAM problems can be
modeled, as shown for a simple single-camera scenario in Figure
\ref{fig:scene_with_block}. It is sufficient to provide the static
priors for the lab-to-world ($\ctrans{\mathrm{w}}{T}{l}$),
tag-2-to-lab ($\ctrans{l}{T}{2}$), tag-105-to-block
($\ctrans{\mathrm{b}}{T}{105}$), and camera-to-rig transform
($\ctrans{\mathrm{r}}{T}{\mathrm{c}}$).  The remaining poses
$\ctrans{\mathrm{w}}{T}{\mathrm{b}}(t)$ and
$\ctrans{\mathrm{w}}{T}{\mathrm{r}}(t)$, as well as the missing poses
of the tags on the block can be determined from the images arriving at
the camera.

\section{Factor Graph}
Our SLAM is formulated as a bipartite factor graph with two types of nodes:
the variables (poses) which are elements of the set $\Theta$, and the
factors, which constrain the variables via the set of measurements  $\mathcal{Z}$. The
factor graph defines a probability $P(\Theta|\mathcal{Z})$ that
assumes its maximum a posteriori (MAP) value for the optimal variable set
$\Theta^*$, given the measurements:
\begin{equation}
\Theta^* = \argmax_\Theta P(\Theta|\mathcal{Z})\ .
\end{equation}
The set of variables $\mathcal{Z}$ contains:
\begin{itemize}
\item The camera poses $\ctrans{\mathrm{body}(\mathrm{cam}\ j)}{T}{\mathrm{cam}\ j}$, with respect to the bodies they are attached to.
\item The tag poses $\ctrans{\mathrm{body}(\mathrm{tag}\ k)}{T}{\mathrm{tag}\ k}$,  relative to their respective bodies.
\item The world poses $\ctrans{\mathrm{w}}{T}{\mathrm{body}\ l}$, of static bodies.
\item The time-dependent world poses $\ctrans{\mathrm{w}}{T}{\mathrm{body}\ m}(t)$, $t = 1\dots N_{\mathrm{t}}$ of dynamic bodies.
\end{itemize}
The likelihood is expressed \cite{kaess2011} as a product of factors $p^{(i)}$
that connects the variables with each other via measurements to form the desired graph structure:
\begin{equation}
  \label{eq:totalprob}
  P(\Theta|\mathcal{Z}) = \prod_ip^{(i)}(\Theta|\mathcal{Z})\ .
\end{equation}
To make the factors $p^{(i)}$ computationally tractable, we follow the standard
approach \cite{kaess2011} and model them as Gaussians:
\begin{equation}
  g(x;\mu,\Sigma) = \exp(-\frac{1}{2}||x\ominus\mu||_\Sigma^2)\ .
\end{equation}
Here, $x$ is the variable, $\mu$ the center of the Gaussian, and $\Sigma$ defines the Mahalanobis distance.
Note the use of the $\ominus$ operator, which reduces to straight subtraction for elements of a vector space,
but produces 6-dimensional Lie algebra coordinates when applied to elements on the SE(3) manifold:
\begin{equation}
  \label{eq:ominus}
  \ctrans{}{T}{A} \ominus \ctrans{}{T}{B} =
  \left[[\log(\mathrm{Rot}(\ctransi{}{T}{B}\ctrans{}{T}{A}))]_\vee^\top,\mathrm{Trans}(\ctransi{}{T}{B}\ctrans{}{T}{A})^\top\right]^\top\ .
\end{equation}
In (\ref{eq:ominus}) Rot() and Trans() refer to the rotational and translational part of the SE(3) transform,
respectively, log() is the matrix logarithm, and $\vee$ denotes the
{\em vee} map operator. Equipped with the definition of a Gaussian
on SE(3), we can now introduce the basic factors $p^{(i)}$ from Eq.\ (\ref{eq:totalprob}).

{\em Absolute Pose Prior}. This unary factor can be used to specify a prior pose
$\ctrans{}{T}{0}$ with noise $\Sigma$ for e.g. a tag or a camera:
\begin{equation}
  p_A(\ctrans{}{T}{}|\ctrans{}{T}{0},\Sigma) = g(\ctrans{}{T}{}; \ctrans{}{T}{0}, \Sigma)\ .
\end{equation}

{\em Relative Pose Prior}. With this binary factor, a known transform $\Delta \ctrans{}{T}{}$
between two pose variables can be specified, with noise $\Sigma$:
\begin{equation}
  p_R(\ctrans{}{T}{A}, \ctrans{}{T}{B}|\Delta\ctrans{}{T}{},\Sigma) = g(\ctransi{}{T}{B}\ctrans{}{T}{A}; \Delta\ctrans{}{T}{}, \Sigma)\ .
\end{equation}
If odometry body pose differences $\Delta \ctrans{}{T}{\mathrm{odom}}(t)$ with noise $\sigma$
are available from e.g. a VIO algorithm running
alongside TagSLAM, a relative pose prior of
$p_R(\ctrans{\mathrm{w}}{T}{\mathrm{body}}(t),
\ctrans{\mathrm{w}}{T}{\mathrm{body}}(t-1)|
\Delta \ctrans{}{T}{\mathrm{odom}}(t),\sigma)$ can be used to insert
the odometry updates into the pose graph.

{\em Tag Projection Factor}. The output of the tag detection
library is a list of tag IDs and the corresponding image
coordinates $\mv{u}_c$ (in units of pixels) of the corners
$c=1\dots 4$ of every tag. This gives rise to one quaternary
tag projection factor per tag:
\begin{equation}
  \begin{split}
    p_T(\ctrans{\mathrm{w}}{T}{\mathrm{body}},
    \ctrans{\mathrm{rig}}{T}{\mathrm{cam}},
    \ctrans{\mathrm{body}}{T}{\mathrm{tag}},
    \ctrans{\mathrm{w}}{T}{\mathrm{rig}}|\{\mv{u}_c\},\sigma_p) =\\
    \prod_{c=1\dots 4}g(\Pi(\ctrans{\mathrm{cam}}{T}{\mathrm{rig}}
    \ctrans{\mathrm{rig}}{T}{\mathrm{w}}
    \ctrans{\mathrm{w}}{T}{\mathrm{body}}
    \ctrans{\mathrm{body}}{T}{\mathrm{tag}}\mv{s}_c);
    \mv{u}_c,\sigma_p)\ .
  \end{split}
\end{equation}
Here, $\mv{s}$ refers to the corner coordinates in the tag reference frame,
i.e. $\mv{s}_1 = [-l/2, -l/2, 0]^\top$,
$\mv{s}_2 = [l/2, -l/2, 0]^\top$,
$\mv{s}_3 = [l/2,  l/2, 0]^\top$,
$\mv{s}_4 = [-l/2, l/2, 0]^\top$
for a tag of side length $l$. A sequence of transforms expresses $\mv{s}$ in
camera coordinates, after which the function $\Pi$ projects \cite{ma2003} the point onto the
sensor plane and converts it to pixel coordinates. The noise parameter $\sigma_p$
is a diagonal matrix that reflects the accuracy of the tag library's corner detector,
which is usually assumed to be about one pixel.

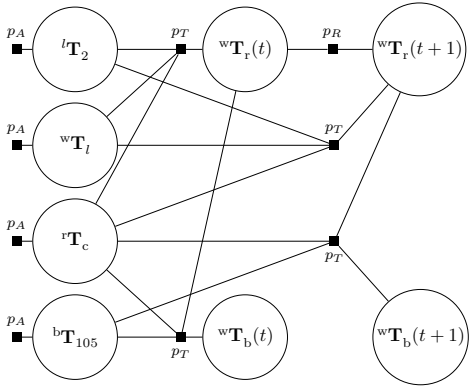
\begin{figure}[ht]
  \vspace{0.25cm}
  \newcommand{\relfacgraphsize}{0.75}
  \begin{center}
    \begin{tabular}{cc}
      \resizebox{\relfacgraphsize\columnwidth}{!}{
      \begin{tikzpicture}

\newcommand{\ysepstatic}{0.2cm}  % y sep static poses
\newcommand{\xsepprior}{0.2cm}   % x sep of priors
\newcommand{\xsepdyn}{1.6cm}     % x sep dyn pose
\newcommand{\xsepfac}{1.1cm}     % x sep dyn pose
\newcommand{\nsize}{16mm}        % minimum node size
\newcommand{\facoff}{12mm}       % factor offset
\newcommand{\xsepfactpo}{4.0cm}  % x sep dyn pose t+1
\newcommand{\xseprelfac}{0.75cm} % rel pose prior factor
\tikzstyle{node} = [latent,minimum size=\nsize];

%
%
% The static poses
%
\node[node]                               (lab_T_2)     {$\ctrans{l}{T}{2}$};
\node[node, below=\ysepstatic of lab_T_2]   (w_T_lab)     {$\ctrans{\mathrm{w}}{T}{l}$};
\node[node, below=\ysepstatic of w_T_lab]   (rig_T_cam)   {$\ctrans{\mathrm{r}}{T}{\mathrm{c}}$};
\node[node, below=\ysepstatic of rig_T_cam] (block_T_105) {$\ctrans{\mathrm{b}}{T}{105}$};

%
%
% priors on static poses
%
\factor[left=\xsepprior of lab_T_2]     {pAtag2} {above:{$p_A$}} {lab_T_2} {};
\factor[left=\xsepprior of w_T_lab]     {pAlab}  {above:{$p_A$}} {w_T_lab} {};
\factor[left=\xsepprior of rig_T_cam]   {pAcam}  {above:{$p_A$}} {rig_T_cam} {};
\factor[left=\xsepprior of block_T_105] {pA105}  {above:{$p_A$}} {block_T_105} {};

%
% dynamic poses at t
%
\node[node,right=\xsepdyn of lab_T_2]     (w_T_rig_t)   {$\ctrans{\mathrm{w}}{T}{\mathrm{r}}(t)$};
\node[node,right=\xsepdyn of block_T_105] (w_T_block_t) {$\ctrans{\mathrm{w}}{T}{\mathrm{b}}(t)$};

%
% tag projection factors at t
%
\factor[right=\xsepfac of lab_T_2] {pT2t}       {above:{$p_T$}} {lab_T_2,w_T_lab,rig_T_cam,w_T_rig_t}{};
\factor[right=\xsepfac of block_T_105] {pT105t} {below:{$p_T$}} {block_T_105,w_T_rig_t,rig_T_cam,w_T_block_t}{};

%
% dynamic poses at t+1
%
\node[node,right=\xsepdyn of w_T_rig_t]   (w_T_rig_tp1)   {$\ctrans{\mathrm{w}}{T}{\mathrm{r}}(t+1)$};
\node[node,right=\xsepdyn of w_T_block_t] (w_T_block_tp1) {$\ctrans{\mathrm{w}}{T}{\mathrm{b}}(t+1)$};

%
% tag projection factors at t + 1
%
\factor[right=\xsepfactpo of w_T_lab]   {pT2tp1}  {above:{$p_T$}} {lab_T_2,w_T_lab,rig_T_cam,w_T_rig_tp1}{};
\factor[right=\xsepfactpo of rig_T_cam] {pT105tp1}{below:{$p_T$}} {block_T_105,w_T_rig_tp1,rig_T_cam,w_T_block_tp1}{};

%
% relative pose priors
%

\factor[right=\xseprelfac of w_T_rig_t]  {prelrig}{above:{$p_R$}} {w_T_rig_t,w_T_rig_tp1}{};

\end{tikzpicture}
      }
    \end{tabular}
  \end{center}
  \caption{Factor graph of Eq.\ \ref{eq:totalprob} for the scene shown in Fig.\ \ref{fig:scene_with_block}}
  \label{fig:sample_graph}
\end{figure}

We can visualize the factor structure of Eq.\ (\ref{eq:totalprob}) by means of
a graph as shown in Fig.\ \ref{fig:sample_graph} for the scene from
Fig.\ \ref{fig:scene_with_block}. In Fig.\ \ref{fig:sample_graph},
black squares represent factors, whereas circles denote pose variables
to be optimized.
The prior factors $p_A$ constrain
the static poses  $\ctrans{l}{T}{2}$, $\ctrans{\mathrm{w}}{T}{l}$,
$\ctrans{\mathrm{r}}{T}{\mathrm{c}}$, and
$\ctrans{\mathrm{b}}{T}{105}$. A tag projection factors $p_T$ arising
from an observation of Tag 2 determines the dynamic rig pose
$\ctrans{\mathrm{w}}{T}{\mathrm{r}}(t)$, whereas an observation of Tag
105 likewise yields the block pose
$\ctrans{\mathrm{w}}{T}{\mathrm{b}}(t)$. Assuming that odometry for
the rig is provided by some external algorithm, there is a relative
pose factor $p_R$ connecting the rig poses for $t$ and $t+1$. Two more
tag observations at $t+1$ generate additional factors that further constrain rig and block poses at $t+1$.

\section{Robust Initialization}
\label{sec:robustinit}
Nonlinear non-convex optimizers such as GTSAM \cite{kaess2011} are
iterative solvers and hence rely on a starting guess that is
reasonably close to the optimum. If for example a camera pose is
initialized such that one of the observed tag corner points lies
behind the camera, the optimizer will likely fail.

Several sources of error can contribute to poor pose
initialization. TagSLAM has been used for several projects already,
and in our experience, human errors are frequently the root cause,
such as inaccuracies or outright typos when entering measured tag
poses, intrinsic or extrinsic calibration errors, duplicate tag ids,
errors in tag size due to printing or misspecfication, using tags that
are not planar, or supplying unsychronized stereo images. One of the
design goals of TagSLAM was to produce, as much as possible, a
reasonable result with a bounded error even with compromised input
data, such that at least the scene can be visualized for further
analysis.

Even when all input data is correct, initializing a camera or tag pose
can be challenging, for example because the corner of the tag is not
detected accurately, which may happen under motion blur, or when a tag
is partially occluded.  As described in Ref.\ \cite{schweighofer2006}, there
are two camera poses from where a single tag looks quite similar, with
only perspective distortion distinguishing between them. For a tag
that is barely large enough to be detected and is viewed at a shallow
angle, a tag corner error of just a single pixel can lead to a
dramatically different pose initialization. Localizing off of a single
tag is therefore intrinsically difficult, and should be
avoided. Satisfactory results from a single tag can only be expected
in combination with odometry input or when the tag image is
sufficiently large.

When multiple cameras and several tags are involved with known or
unknown poses it is anything but obvious which measurements to use,
and in which order.  For example, should the pose of camera 1 be
established first from the tag corners, then the pose of camera 2 via
a known extrinsic calibration, or the other way round? Which tags
should be used for this purpose if several are visible?

In case the tag poses are known, one might be tempted to answer the
last question with: use the corner points of all tags simultaneously
with a perspective n-point method (PnP). In
practice we find that PnP is not robust to misspecified tag
poses. Moreover when it fails it is not clear which of the tags is in
error, necessitating an expensive elimination process. For this reason
we base all pose initialization on homographies \cite{ma2003} from a
minimum set of tags only, carefully picking which tags to use, and in
which order. The remainder of this section will describe how exactly
this is done.

To fully exploit the history of observations, two separate graphs are
maintained: the {\em full graph} contains factors and variables
pertaining to all measurements up to the current time, whereas the
{\em optimized graph} only has those variables and factors that are
sufficiently constrained to form a well conditioned optimization
problem. All incoming measurements are thus entered immediately into
the {\em full graph}, but only find their way into the {\em optimized
  graph} when the variables can actually be initialized.

Usually several static poses can be initialized right away because a
prior is available. In Fig.\ \ref {fig:sample_graph} for example, the
poses $\ctrans{l}{T}{2}$, $\ctrans{\mathrm{w}}{T}{l}$,
$\ctrans{\mathrm{r}}{T}{\mathrm{c}}$, and
$\ctrans{\mathrm{b}}{T}{105}$ are determinable due to the prior
factors shown to their left, and can therefore be directly inserted
into the {\em optimized graph}.

\subsection{Subgraph discovery}
As measurements arrive, they give rise to new factors that create
edges in the graph between existing and new variables. Variables
connected to these factors may be rendered determinable, which in
turn, through existing factors, may cause other variables to become
determinable as well. Thus every new factor can give rise to a
subgraph of newly determinable variables. We refer as {\em subgraph
  discovery} to the exhaustive traversing of the graph until no more
new variables can be determined. During this process, all
deteriminable variables and factors are entered into the
subgraph. Further, an {\em initialization list} is created that
contains the factors in the order they are discovered. The
initialization list later governs the order in which the subgraph will
be initialized.  The discovered subgraph depends on the new factor
from which the discovery is started, so potentially, each new factor
gives rise to a different subgraph. In practice the subgraphs are
connected, and often a set of new factors generates only a single
subgraph.

Note that also factors arising from past measurements may enter the
subgraph. For example, if only tags $A$ and $B$ with unknown pose were
seen at time step $t$, no camera rig pose could be determined. But if
at $t+1$ tags $B$ and $C$ are observed, and tag $C$ has known pose,
then the subgraph at $t+1$ will contain poses for tags $A$, $B$, and
$C$, although tag $A$ was observed in the previous step.

For robustness against ambiguity, a tag pose will not enter a
subgraph unless its pose is either already determined from previous
observations, or given by a pose prior, or has a pose ambiguity error
ratio \cite{munoz2018} of $e(\gamma)/e(\dot{\gamma}) > 0.3$ while
being viewed at an angle of less than 60$^{\circ}$. The last condition
reflects the fact that tag pose ambiguity is most serious a problem
when the tag is seen at a small angle.

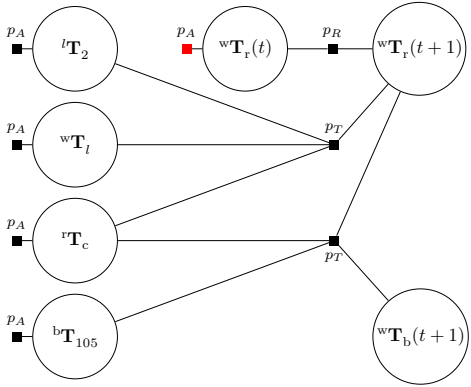
\begin{figure}[ht]
  \newcommand{\relfacgraphsize}{0.75}
  \begin{center}
    \begin{tabular}{cc}
      \resizebox{\relfacgraphsize\columnwidth}{!}{
      \begin{tikzpicture}

\newcommand{\ysepstatic}{0.2cm}     % y sep static poses
\newcommand{\xsepprior}{0.2cm}      % x sep of priors
\newcommand{\xsepdyn}{1.6cm}        % x sep dyn pose
\newcommand{\xsepfac}{1.1cm}        % x sep dyn pose
\newcommand{\nsize}{16mm}        % minimum node size
\newcommand{\facoff}{12mm}        % factor offset
\newcommand{\xsepfactpo}{4.0cm}        % x sep dyn pose t+1
\newcommand{\xseprelfac}{0.75cm}        % rel pose prior factor
\tikzstyle{node} = [latent,minimum size=\nsize];

%
%
% The static poses
%
\node[node]                               (lab_T_2)     {$\ctrans{l}{T}{2}$};
\node[node, below=\ysepstatic of lab_T_2]   (w_T_lab)     {$\ctrans{\mathrm{w}}{T}{l}$};
\node[node, below=\ysepstatic of w_T_lab]   (rig_T_cam)   {$\ctrans{\mathrm{r}}{T}{\mathrm{c}}$};
\node[node, below=\ysepstatic of rig_T_cam] (block_T_105) {$\ctrans{\mathrm{b}}{T}{105}$};

%
%
% priors on static poses
%
\factor[left=\xsepprior of lab_T_2]     {pAtag2} {above:{$p_A$}} {lab_T_2} {};
\factor[left=\xsepprior of w_T_lab]     {pAlab}  {above:{$p_A$}} {w_T_lab} {};
\factor[left=\xsepprior of rig_T_cam]   {pAcam}  {above:{$p_A$}} {rig_T_cam} {};
\factor[left=\xsepprior of block_T_105] {pA105}  {above:{$p_A$}} {block_T_105} {};

%
% dynamic poses at t
%
\node[node,right=\xsepdyn of lab_T_2]     (w_T_rig_t)   {$\ctrans{\mathrm{w}}{T}{\mathrm{r}}(t)$};
\factor[color=red,left=\xsepprior of w_T_rig_t]  {pwTrigt}  {above:{$p_A$}} {w_T_rig_t} {};

%
% dynamic poses at t+1
%
\node[node,right=\xsepdyn of w_T_rig_t]   (w_T_rig_tp1)   {$\ctrans{\mathrm{w}}{T}{\mathrm{r}}(t+1)$};
\node[node,right=\xsepdyn of w_T_block_t] (w_T_block_tp1) {$\ctrans{\mathrm{w}}{T}{\mathrm{b}}(t+1)$};

%
% tag projection factors at t + 1
%
\factor[right=\xsepfactpo of w_T_lab]   {pT2tp1}  {above:{$p_T$}} {lab_T_2,w_T_lab,rig_T_cam,w_T_rig_tp1}{};
\factor[right=\xsepfactpo of rig_T_cam] {pT105tp1}{below:{$p_T$}} {block_T_105,w_T_rig_tp1,rig_T_cam,w_T_block_tp1}{};

%
% relative pose priors
%

\factor[right=\xseprelfac of w_T_rig_t]  {prelrig}{above:{$p_R$}} {w_T_rig_t,w_T_rig_tp1}{};

%\factor[right=\xseprelfac of w_T_block_t] {prelblock}{above:{$p_R$}} {w_T_block_t,w_T_block_tp1}{};

\end{tikzpicture}
      }
    \end{tabular}
  \end{center}
  \caption{Subgraph of graph in Fig.\ \ref{fig:sample_graph}}
  \label{fig:sample_subgraph}
\end{figure}

All variables of the so generated subgraph are determinable, but if
any dynamic poses from previous time steps are present, they must be
constrained or the problem is ill determined. This is done by
inserting absolute priors for those
poses. Fig.\ \ref{fig:sample_subgraph} shows a subgraph, derived from
the graph in Fig.\ \ref{fig:sample_graph} during time step $t+1$. Note
the removal of the tag projection factors and of the block pose for
time $t$, and the addition of a prior (colored red) on pose
$\ctrans{\mathrm{w}}{T}{\mathrm{r}}(t)$.

To arrive at a complete set of subgraphs, all new factors are entered
into the {\em new factor list}. The order of the factors is important,
and will be discussed further below. An iteration over this list is
performed, and, unless the new factor is already part of an already
discovered subgraph, a new subgraph is generated by discovery. This
procedure results in a set of disjoint subgraphs with all determinable
variables and connected factors, as well as the corresponding {\em
  initialization lists}.

\subsection{Order of subgraph discovery}

What still remains to be settled is the order in which factors are
entered into the {\em new factor list}. This strongly affects the {\em
  initialization lists} generated during subgraph discovery, and hence
the order in which measurements are used for initialization. For
instance in Fig.\ \ref{fig:sample_subgraph}, there are two ways to
initialize $\ctrans{\mathrm{w}}{T}{\mathrm{r}}(t+1)$. One is through
the relative pose factor with respect to
$\ctrans{\mathrm{w}}{T}{\mathrm{r}}(t)$, the other through the tag
projection factor on tag 2.

By examining the typical failure cases of several alternative
approaches, the following order for entering factors into the {\em new
  factor list} was established:
\begin{enumerate}[wide, labelwidth=!]
\item Any relative pose priors. Note that this prefers odometry
  updates, which are typically more reliable in establishing a body
  pose than initializing it from tag observations.
\item Any tag projection factors. These factors are sorted in
  descending order by pixel area of the observed tag, irrespective
  which camera they were observed from. This means that larger tags
  will be used first to establish a pose.
\item Any factors that do not establish a pose. Examples are problem
  specific additional factors such as distance measurements between
  tag corners.
\end{enumerate}

\subsection{Optimization}
Once the subgraphs have been obtained, their variables are initialized
in the order specified by the corresponding {\em initialization
  list}. Then all subgraphs are optimized using GTSAM, in
non-incremental mode, i.e. without using iSAM2 \cite{kaess2011}, and
their error is evaluated. This step is analogous to model validation
in RANSAC. If a subgraph's error falls below a
configurable threshold, the subgraph is accepted, and its factors and
optimized values are transferred to the {\em optimized
  graph}. Subsequently, the {\em optimized graph} is optimized with an
iSAM2 update step.

In the rare case where a subgraph is rejected due to excessive error,
an initialization with different ordering is attempted. Since
exhaustively trying all possible orderings is computationally too
expensive, an ad hoc procedure is adopted that was found to work
satisfactorily in practice: the {\em initialization list} of the
subgraph is rotated such that the first factor goes to the end of the
list, and all other elements of the list advance by one. This implies
that successively smaller tags are used to seed the initialization
process. The subgraph is initialized with the new ordering, followed
by optimization. In case the subgraph error is still too high, this
process is repeated until the original ordering is reached again. If
no initialization ordering leads to an acceptable error, the subgraph
is rejected, thus preventing an outlier measurement from contaminating
the {\em optimized graph}.

\subsection{Diagnostics}
Rejection of a subgraph is usually a strong indicator of faulty input
parameters or misdetected tag corners. In most cases however, subgraph
rejection is not fatal, and TagSLAM will successfully handle
subsequent incoming data. For diagnosis, TagSLAM produces per-factor
and time resolved error statistics. Such output is essential for
tracking down e.g. incorrectly specified tag poses.

\begin{figure}[t]
  \centering
  \vspace{0.25cm}
  \includegraphics[width=0.65\columnwidth]{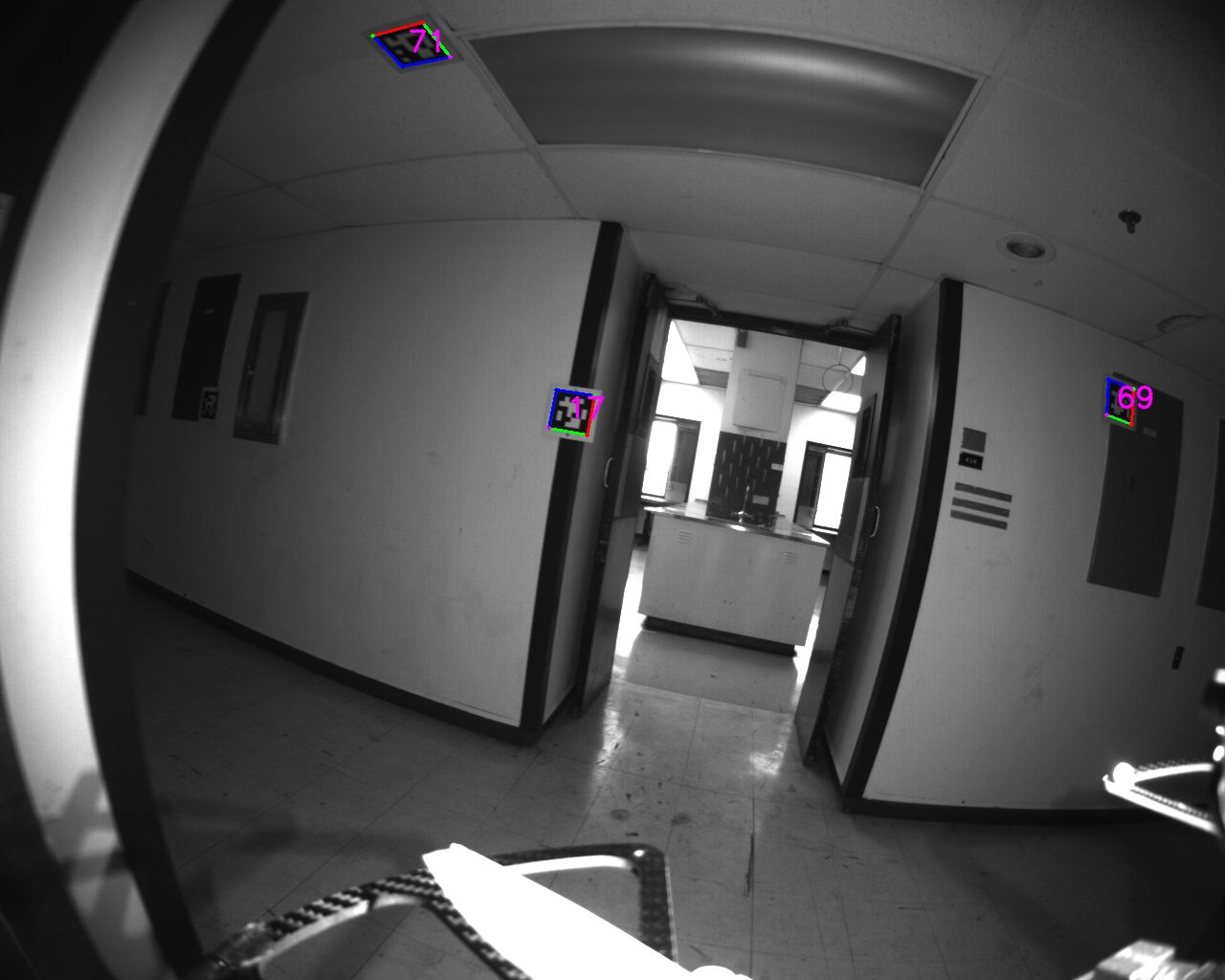}
  \caption{View from the left camera of an Open Vision
    Computer. Detected AprilTags are shown in color.}
  \label{fig:corridor_ovc}
\end{figure}
\section{Application Example}
\label{sec:application}
We illustrate the versatility of TagSLAM by showing how it can be used
to achieve loop closure for VIO. The synchronized images and IMU data
that serve as input for the VIO are collected with an Open Vision
Computer \cite{quigley2018}. During 13 minutes, a total of 15595
stereo frames are recorded at 20Hz along the 630m long trajectory
through the rooms of an abandoned chemical
laboratory. Fig.\ \ref{fig:corridor_ovc} shows an example image of
some of the 57 tags that are strategically placed along the
corridor. Their poses are deterimined from the wall orientations and
from laser distance measurements with a Leica Disto D3a. The odometry
is computed with the stereo VIO algorithm as described in
Ref. \cite{sun2018} but, running offline with abundant CPU resources
available, we use a larger number of visual feature points to improve
drift.

\begin{figure}[ht]
  \centering
  \includegraphics[angle=90, width=0.5\columnwidth]{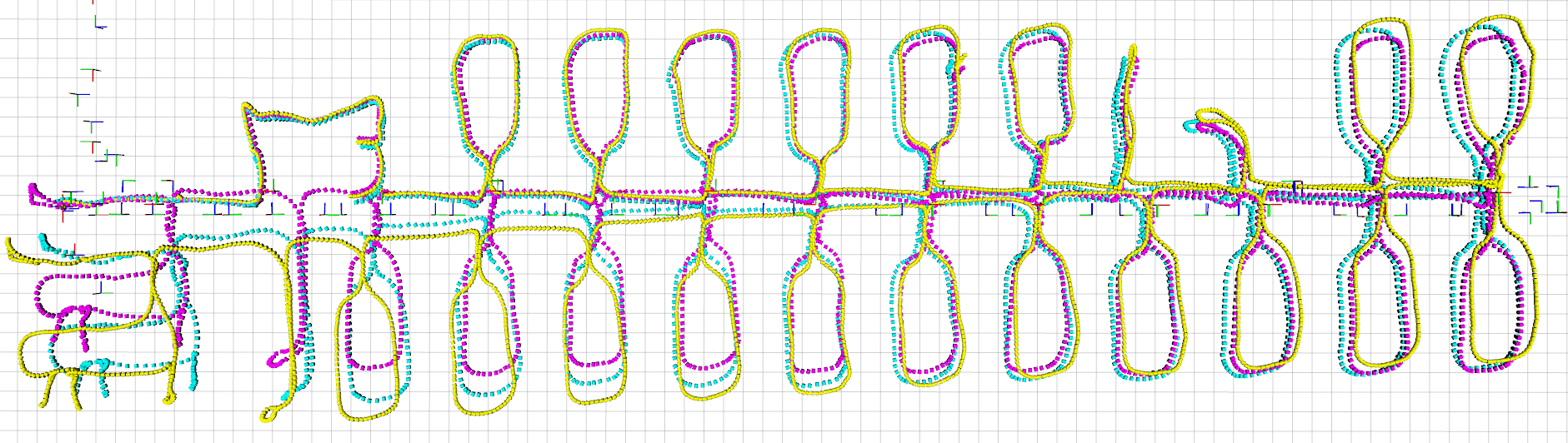}
  \caption{Trajectory using {\color{cyan} VIO}, {\color{magenta} TagSLAM},
    and {\color{yellow} ORB-SLAM2}. The grid cell size is 1m.}
  \label{fig:loop_closure}
\end{figure}

Fig.\ \ref{fig:loop_closure} shows the trajectories for VIO (cyan),
loop-closed TagSLAM (magenta), and stereo ORB-SLAM2
\cite{murartal2017} (yellow). The tag locations are visible in the map
as well. All trajectories start at the same point at the bottom of
the map, but only the TagSLAM trajectory returns correctly to the
starting point. Both VIO and ORB-SLAM2 exhibit drift, and evidently
ORB-SLAM2 does not achieve loop closure. This is not surprising since
the hallway images look very different while returning.  By combining
tag projection factors from the camera images with relative pose
factors from the odometry, TagSLAM by design closes the loop.

Creating and incrementally optimizing the graph upon factor insertion
takes 188s on an Intel i9-9900K 3.6GHz CPU, which is an average
performance of about 12ms per frame. A final full (non-incremental)
graph optimization adds another 4.3s to the total processing
time. While 12ms time per frame seems to indicate the possibility of
running in real time, individual frames can take longer to process,
depending on iSAM2 relinearization. As the graph grows over time, so
does the CPU load, and individual frames can take as much as
260ms. However, for situations where there already is a trusted map of
tag poses available, TagSLAM can be configured to retain only the last
two time steps in the graph, making it suitable for real-time
operation.

\section{Conclusion}
In this paper we present TagSLAM, a highly flexible front-end to the
factor graph optimizer GTSAM that makes fiducial based visual SLAM and
related tasks accessible to the robotics community by means of an open
source ROS package located at
\url{https://berndpfrommer.github.io/tagslam_web}.

\balance
\bibliographystyle{IEEEtran}
\bibliography{IEEEabrv,IEEEconfabrv,root}
\end{document}